\documentclass[10pt,twocolumn,letterpaper]{article}

\usepackage{cvpr}              %
\usepackage{float}
\usepackage{booktabs} %
\usepackage{multirow} %
\usepackage{graphicx} %
\usepackage{caption}
\usepackage{float}
\usepackage{amssymb}
\usepackage{bbm}
\newcommand{\ind}[1]{\mathbbm{1}\{#1\}}
\usepackage{amsthm}

\newcommand{\relu}{\mathrm{ReLU}}

\definecolor{cvprblue}{rgb}{0.21,0.49,0.74}
\usepackage[pagebackref,breaklinks,colorlinks,allcolors=cvprblue]{hyperref}

\def\paperID{5787} %
\def\confName{CVPR}
\def\confYear{2026}

\title{\raisebox{-0.3em}{\includegraphics[height=1.5em]{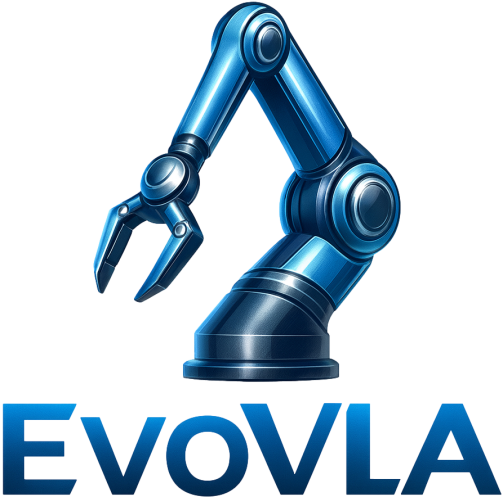}}~EvoVLA: Self-Evolving Vision-Language-Action Model} %

\author{
    \textbf{Zeting Liu}$^{*}$ \quad
    \textbf{Zida Yang}$^{*}$ \quad
    \textbf{Zeyu Zhang}$^{*\dag}$ \quad
    \textbf{Hao Tang}$^{\ddag}$ \vspace{0.1cm}\\
    Peking University\\
    \small $^*$Equal contribution. $^\dag$Project lead.
    $^\ddag$Corresponding author: bjdxtanghao@gmail.com.
}

\begin{document}
\twocolumn[{%
\renewcommand\twocolumn[1][]{#1}%
\maketitle
\vspace{-0.35cm}
\includegraphics[width=\textwidth]{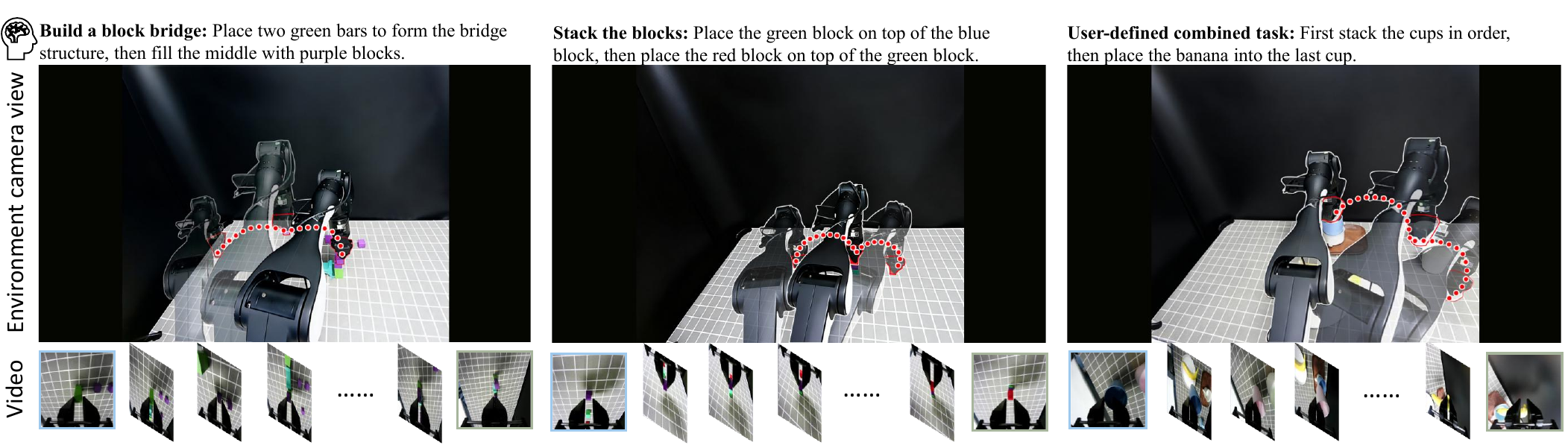}
\captionof{figure}{\textbf{EvoVLA: Addressing stage hallucination in long-horizon manipulation.} Our framework combines Stage-Aligned Reward (SAR), Pose-Based Object Exploration (POE), and Long-Horizon Memory to achieve robust performance on complex manipulation tasks. Example results shown: Block Bridge, Stack, and Cup Stacking with Insertion tasks across simulation and real-world deployment.}\label{fig:teaser}
\vspace{1em} 
}]

\begin{abstract}
Long-horizon robotic manipulation remains challenging for Vision-Language-Action (VLA) models despite recent progress in zero-shot generalization and Sim2Real transfer. Current VLA models suffer from stage hallucination, where agents exploit coarse evaluation signals to shortcut multi-step tasks, reporting high progress without actual task completion. We present EvoVLA, a self-supervised VLA framework addressing this through three synergistic components: Stage-Aligned Reward (SAR) uses triplet contrastive learning with Gemini-generated hard negatives to prevent visual shortcuts; Pose-Based Object Exploration (POE) grounds curiosity in relative object-gripper pose rather than pixels; and Long-Horizon Memory with selective context and gated fusion stabilizes intrinsic shaping. Extensive evaluations on Discoverse-L, a long-horizon manipulation benchmark with three multi-stage tasks, demonstrate that EvoVLA improves average success by 10.2 percentage points over the strongest baseline (OpenVLA-OFT), reaching 69.2\%, achieves 1.5$\times$ better sample efficiency, and reduces stage hallucination from 38.5\% to 14.8\%. Real-world deployment on physical robots achieves 54.6\% average success across four manipulation tasks, outperforming OpenVLA-OFT by 11.0 points, demonstrating effective Sim2Real transfer and robust generalization.
Code: \url{https://github.com/AIGeeksGroup/EvoVLA}.
Website: \url{https://aigeeksgroup.github.io/EvoVLA}.
\end{abstract}

\section{Introduction}\label{sec:intro}

Vision-Language-Action (VLA) models promise generalist robot policies by unifying perception, language, and control within a single large-scale backbone. Early systems such as Do-As-I-Can~\cite{ahn2022can}, Inner Monologue~\cite{huang2022inner}, and Code-as-Policies~\cite{liang2023code} demonstrated that language grounding can guide imitation and reinforcement learning, while recent foundation models, including PaLM-E~\cite{driess2023palm}, RT-X/RT-2~\cite{brohan2022rt,brohan2023rt2}, OpenVLA~\cite{kim2024openvla}, and the $\pi_0$ family~\cite{black2024pi0,openpi2024} with FAST tokenization~\cite{fast2025}, scale this recipe to web-scale corpora and thousands of robot demonstrations, achieving compelling zero-shot behavior on many short-horizon skills. Yet when tasks require dozens of semantically distinct stages and persistent state tracking, current VLAs still rely on brittle waypoints or ad-hoc progress classifiers; without explicit stage semantics and temporal grounding, autonomy often collapses despite strong pretraining.

We identify two long-horizon obstacles. First, reinforcement learning remains sample-inefficient: sparse rewards and high-dimensional observations induce ``stage hallucination'', where the policy farms superficial visual cues to appear successful. Self-supervised reinforcement learning (SSRL) offers dense intrinsic feedback, yet existing curiosity formulations~\cite{pathak2017curiosity,burda2018exploration} treat vision-only novelty detection and quickly collapse in cluttered scenes. Second, long-horizon memory is fragile. Most ``memory-augmented'' VLAs compress history by averaging or truncation, causing catastrophic forgetting. Worse, the community lacks a benchmark that exposes memory drift or enables systematic training—without the right data, improved memory modules cannot be meaningfully evaluated.

These challenges motivate EvoVLA (Figure~\ref{fig:teaser}). To combat hallucination, we introduce a self-supervised reinforcement learning (SSRL) pipeline that couples stage-aware rewards with pose-grounded curiosity, delivering dense signals that respect task semantics. To curb memory forgetting, we take a system view: (i) a Long-Horizon Memory module selects, gates, and writes back only utility-critical context, and (ii) Discoverse-L, a long-horizon manipulation benchmark with ground-truth stage events for evaluation and task-aligned normalization, provides the training/evaluation substrate required to stress-test memory designs.

Our contributions are threefold. First, we present an SSRL framework that fuses stage-aligned rewards, pose-based curiosity, and counterfactual shaping to suppress hallucination while retaining exploration. Second, we systematically tackle long-horizon memory by combining a context-selected gating module with the Discoverse-L benchmark, enabling both training and evaluation of memory-heavy VLAs. Third, we deliver comprehensive experiments across simulation and real robot deployments to validate robustness and sample efficiency.

Hence, we summarize our contributions as follows:

\begin{itemize}
    \item We propose \textbf{EvoVLA}, a self-supervised long-horizon learning method, where SAR combined with Gemini-driven hard negatives and POE provides dense, semantically consistent intrinsic feedback, enabling scalable VLA fine-tuning without additional labels.
    \item We target long-horizon forgetting through a context-selected Long-Horizon Memory mechanism with selective attention and gated fusion, coupled with the \textbf{Discoverse-L} benchmark featuring three multi-stage manipulation tasks (18--74 stages), which together provide the community with both a method and a dataset/measurement suite for studying memory in long-horizon manipulation.
    \item Extensive evaluations show \textbf{EvoVLA} lifts \textbf{Discoverse-L} success to \textbf{69.2\%} (+10.2 points over OpenVLA-OFT~\cite{kim2025openvlaoft}), reaches the 50\% success threshold with 1.5$\times$ fewer environment steps, and attains \textbf{54.6\%} real-robot success (+11.0 points vs.\ OpenVLA-OFT~\cite{kim2025openvlaoft}, +16.9 vs.\ $\pi_0$-FAST~\cite{openpi2024}) while keeping hallucination at \textbf{14.8\%}.
\end{itemize}

\section{Related Work}\label{sec:related}

\noindent\textbf{VLA for Embodied Intelligence.}
Vision-Language-Action (VLA) models replace staged pipelines with end-to-end multimodal learning~\cite{ye2025vla,liu2025nav,song2025hazards,song2025maniplvm,huang20253d,wu2025stereoadapter}. PaLM-E~\cite{driess2023palm} and RT-X/RT-2~\cite{brohan2022rt,brohan2023rt2} scale trajectory pretraining with frozen ViT encoders and action heads, improving generalization and Sim2Real transfer. OpenVLA~\cite{kim2024openvla} and Octo~\cite{octo2024} advance open-source finetuning for real tasks, while RoboMamba~\cite{liu2024robomamba} introduces SSM-based architectures for long sequences. Recent work on optimized fine-tuning~\cite{kim2025openvlaoft} explores parallel decoding, action chunking, and continuous representations to boost VLA efficiency and performance. Yet two issues remain: stage hallucination (exploiting Vision-Language Model (VLM) scores without actual completion) and reward modeling dilemmas in high-dimensional POMDPs. EvoVLA addresses both via stage-aligned rewards and pose-grounded SSRL.

\noindent\textbf{RL for VLA Fine-Tuning and Exploration in Embodied Tasks.}
Reinforcement learning (RL) improves VLA finetuning for long-horizon tasks. PPO~\cite{schulman2017proximal} and SAC~\cite{haarnoja2018soft} optimize action heads but depend on careful reward design. RT-1/RT-2~\cite{brohan2022rt,brohan2023rt2} combine policy optimization with large demonstration corpora, yet remain limited by sparse reward brittleness and hallucination. Recent 2025 VLA RL fine-tuning works explore consistency policies, chunked offline regimes, temporal feedback, and verified rewards~\cite{hu2025flare,chen2025conrft,huang2025corft,shu2025rftf,li2025vlarft,lu2025vlarl,guo2025improving,zhang2025reinbot,li2025simplevla}.
Traditional RL suffers from sparse signals and annotation costs, causing hallucinations and instability. Self-supervised reinforcement learning (SSRL) adds dense intrinsic feedback for efficiency. ICM~\cite{pathak2017curiosity} introduces intrinsic rewards via forward dynamics prediction, Plan2Explore~\cite{sekar2020planning} plans curiosity via trajectory ensembles, and RaMP~\cite{chen2023self} designs random features for zero-shot world-model transfer. Recent VLA-oriented SSRL~\cite{du2021curious,liang2023alp} jointly learns policies and representations, yet pixel-based curiosity remains brittle in cluttered scenes. EvoVLA grounds exploration in geometric pose structure rather than in pixels.

\begin{figure}[t]
    \centering
    \includegraphics[width=\linewidth]{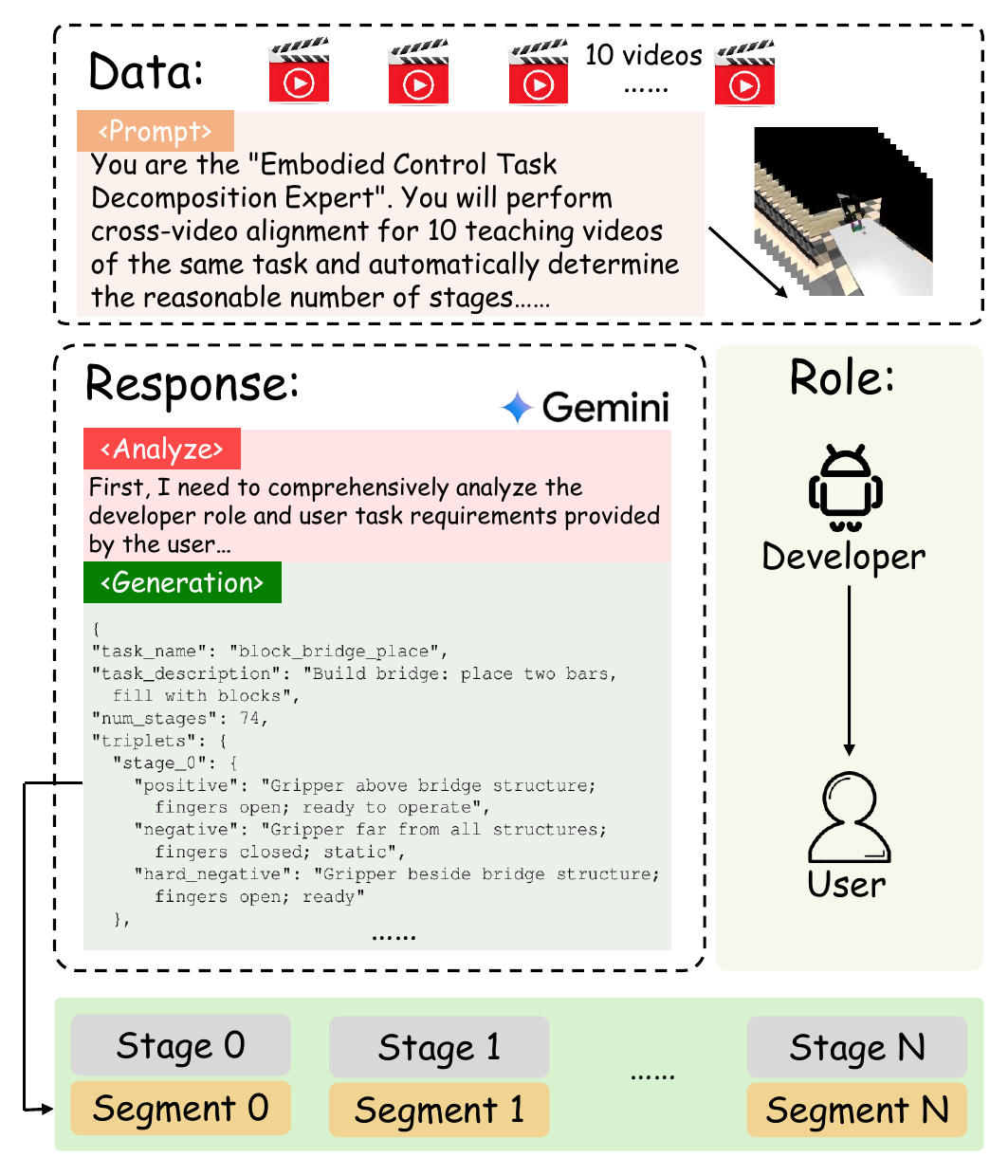}
    \caption{EvoVLA Data Engine. Aligned with Discoverse-L and the video-driven stage discovery pipeline to close the data–reward–policy loop.}\label{fig:data-pipeline}
    \vspace{-0.4cm}
\end{figure}

\noindent\textbf{Counterfactual Shaping for Enhanced Reliability.}
Potential-based reward shaping~\cite{ng1999policy,harutyunyan2015expressing} mitigates reward bias and stage hallucination by adjusting returns without altering optimal policies. Reflexion~\cite{shinn2023reflexion} adds language introspection for long-horizon computation. Memory-augmented systems such as JARVIS-1~\cite{wang2024jarvis} enable open-world multi-task planning; yet, they often rely on non-end-to-end frameworks or short memory windows. EvoVLA bridges this gap by extending counterfactual shaping via SSRL post-training and fusing it with geometric pose-based exploration for data-efficient learning.

\begin{figure*}[!t]
    \centering
    \includegraphics[width=\linewidth]{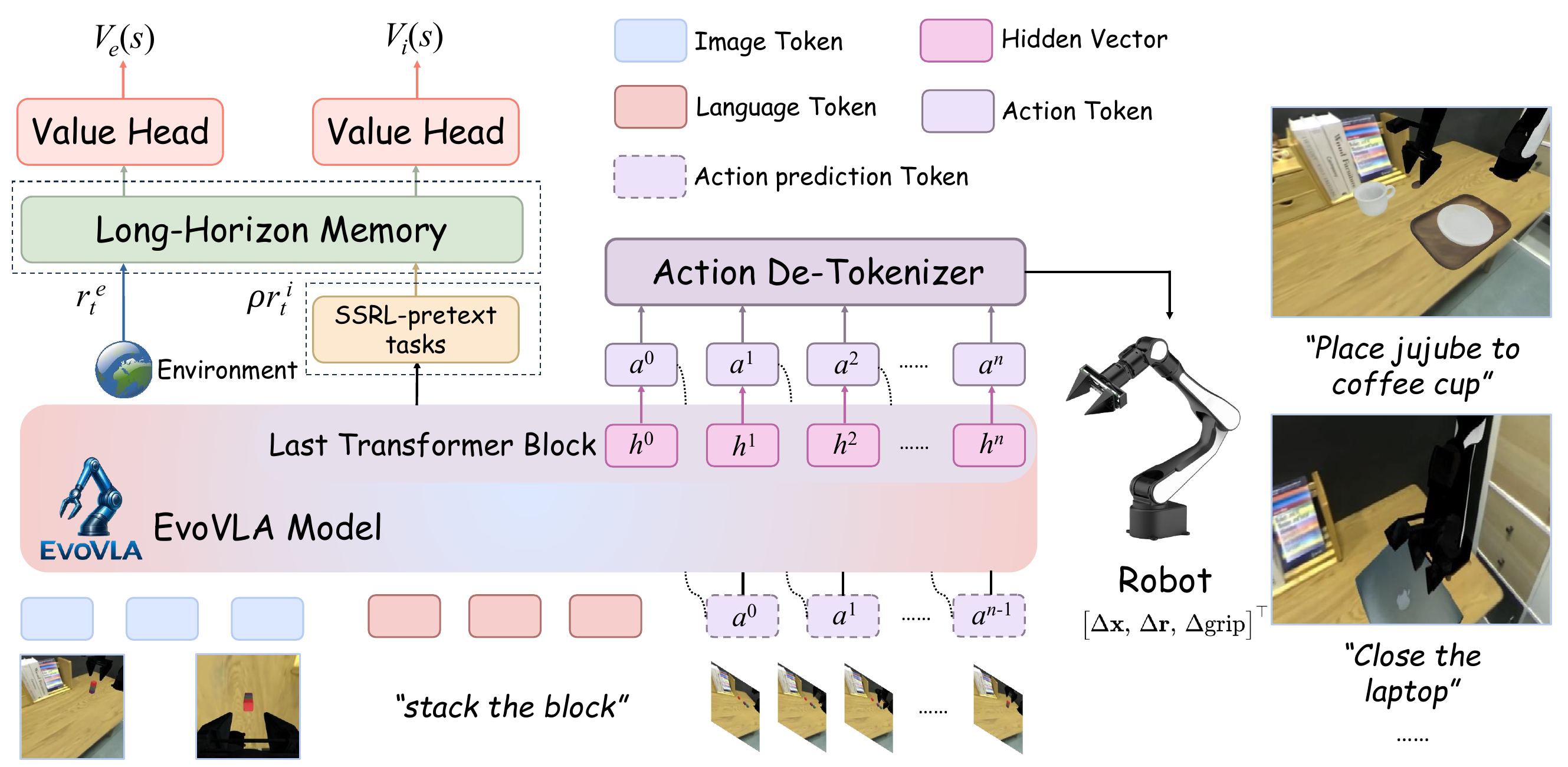}
    \caption{\textbf{EvoVLA overview.} Built on OpenVLA-OFT backbone, EvoVLA integrates three modules: Stage-Aligned Reward (SAR) with hard negatives and temporal smoothing, Pose-Based Object Exploration (POE) via world models, and Long-Horizon Memory with context selection and gated fusion. The framework couples with Discoverse-L for training and deploys to real robots.}%
    \label{fig:overview}
        \vspace{-0.4cm}
\end{figure*}

\section{The Proposed Method}

\subsection{Overview}

EvoVLA couples a task-aligned data pipeline (Figure~\ref{fig:data-pipeline}) with long-horizon control modules (Figure~\ref{fig:overview}). Discoverse-L supplies three AIRBOT-Play tasks with 50 scripted rollouts each; we store trajectories in RLDS~\cite{agarwal2022rlds}, compute task-aligned normalization, and run a multi-step prompting workflow with Gemini~2.5~Pro~\cite{gemini2024api} to produce unified stage dictionaries of positive, negative, and hard-negative predicates (template, validation heuristics, and failure handling receive detailed coverage in Appendix~\ref{app:templates}). The same semantics supervise simulation training and Sim2Real deployment, where policies execute from language prompts; for unseen tasks we regenerate the dictionary from 50 teleoperated demonstrations.

The architecture centers on three interacting modules. SAR scores image-text triplets with CLIP~\cite{radford2021clip} contrast and temporal smoothing, delivering stage-faithful dense rewards. POE grounds curiosity in relative gripper-to-object poses via paired forward/inverse world models to reduce spurious novelty from visual distractors. Long-Horizon Memory selects salient history tokens via attention-based context selection and gates intrinsic shaping so progress signals remain stable. All modules share the OpenVLA-OFT backbone and are optimized jointly via PPO.

Training follows a modular recipe: optionally warm-start with supervised learning on the rollouts, then apply PPO~\cite{schulman2017proximal} with task-aligned normalization while SAR, POE, and memory remain active. The following subsections detail each component.

\subsection{Self-Supervised Reinforcement Learning}

Our self-supervised learning strategy enables the agent to learn efficiently without manual reward engineering. Following prior work on curiosity-driven exploration~\cite{pathak2017curiosity,burda2018exploration}, we generate intrinsic rewards from pretext tasks. However, we ground these signals in stage-aligned evaluation and geometric structure rather than raw pixels, yielding more reliable learning signals. The combined reward is:
\begin{equation}
\label{eq:combined-reward}
\tilde{r}_t = r^e_t + \rho \big(r^{\text{stage}}_t + r^{\text{cur}}_t + r^{\text{prog}}_t\big).
\end{equation}
where $r^e_t$ is the sparse external task reward, $r^{\text{stage}}_t$ is the stage-aligned reward, $r^{\text{cur}}_t$ stems from POE, $r^{\text{prog}}_t$ comes from Long-Horizon Memory, and $\rho$ is the fixed intrinsic weight (we use $\rho=0.6$, selected via sensitivity analysis; Supp. Mat. Fig.~\ref{fig:rho-sensitivity}). Below, we detail each component.

\noindent\textbf{Stage-Aligned Reward.}\label{sec:sar}
We evaluate stage completion through image-text contrastive learning, producing dense reward signals to guide policy learning. The key insight is that standard positive-negative contrastive learning is insufficient, so we need hard negatives that capture near-miss states to prevent policies from exploiting visual shortcuts.

\textit{\underline{Triplet text generation.}}
For each stage $k$, the Gemini~2.5~Pro~\cite{gemini2024api} stage dictionary provides a triplet: positive $T^+_k$ (completion state), mutually-exclusive negative $T^-_k$, and counterfactual hard negative $T^{h-}_k$ (near-miss). The hard negative uses spatial/contact predicates (e.g., ``grasping non-target object'') rather than visual features (e.g., ``grasping blue cup''), ensuring robustness to appearance variations.

\textit{\underline{Image-text contrastive scoring.}}
Given VLM encoder $\phi$ (CLIP~\cite{radford2021clip,huang20253dcoca,huang2025dc}), we compute stage-aligned scores by comparing image observations against the triplet text descriptions:
\begin{align}
\label{eq:sar}
u_k(t) &= \sigma\Big(\tau\,[s^+_k(t) - \max\{s^-_k(t), s^h_k(t)\}]\Big), \\
u_k(t) &\in [0,1], \nonumber
\end{align}
where $s^+_k(t) = \langle \phi_{\text{img}}(o_t), \phi_{\text{text}}(T^+_k) \rangle$ (similarly for $s^-_k(t), s^h_k(t)$), $\tau$ is the temperature parameter, and $\sigma$ is the sigmoid function. This contrastive formulation forces the VLM to distinguish between near-miss states and actual completion.

\textit{\underline{Temporal smoothing.}}
To stabilize score fluctuations and prevent noisy reward signals, we maintain a running average $\bar{u}_k$ that is updated smoothly over time:
\begin{equation}
\bar{u}_k(t) = (1-\alpha) \bar{u}_k(t-1) + \alpha u_k(t).
\end{equation}
where $\alpha$ is the smoothing coefficient. The stage reward is then computed as the progress in these smoothed scores:
\begin{equation}
\label{eq:rstage}
r^{\text{stage}}_t = \bar{u}_{\kappa_t}(t) - \bar{u}_{\kappa_t}(t-1).
\end{equation}
where $\kappa_t$ is the current active stage. We advance to stage $\kappa_t+1$ when a sliding window (size $m=8$) of recent $r^{\text{stage}}$ values exceeds a threshold, indicating sustained progress in the current stage (Appendix~\ref{app:implementation}). This approach filters out transient score fluctuations while preventing premature transitions.

\noindent\textbf{Pose-Based Object Exploration.}\label{sec:poe}
Pixel-based curiosity suffers from spurious novelty, with exploration driven by irrelevant changes, such as lighting variations or camera motion, rather than task progress. POE addresses this by grounding curiosity in task-relevant interaction dynamics between the object and the parallel-jaw gripper, focusing on how the gripper’s motion and contact evolve during manipulation.

\textit{\underline{Latent manipulation dynamics.}}
We represent the manipulation state as $z_t = \psi(T_{\text{ee}}^{-1} T_{\text{obj}}) \in \mathbb{R}^6$, where $T_{\text{ee}}$ is the end-effector pose, $T_{\text{obj}}$ is the object pose, and $\psi$ converts the relative transformation to a 6D vector in axis-angle plus translation representation (3D translation + 3D axis-angle rotation). We train forward and inverse models:
\begin{align}
\label{eq:world-models}
\hat{z}_{t+1} &= f_\phi(z_t, a_t),  \\
\hat{a}_t &= g_\psi(z_t, z_{t+1}).
\end{align}
These lightweight MLP models ($2\times256$ units) learn the dynamics of object manipulation without being distracted by visual nuisances.

\textit{\underline{Task-relevant intrinsic rewards.}}
The curiosity reward encourages exploring states where the forward model makes large prediction errors, while the progress reward captures learning improvements:
\begin{align}
r^{\text{cur}}_t &= \frac{\eta}{2} \|\text{sg}(\hat{z}_{t+1}) - z_{t+1}\|_2^2, \\
r^{\text{base}}_t &= \relu\!\big(\overline{\mathcal{L}_F}(t-1) - \overline{\mathcal{L}_F}(t)\big).
\end{align}
where $\text{sg}(\cdot)$ denotes stop-gradient (preventing policy from exploiting model errors), $\overline{\mathcal{L}_F}$ is the forward loss with temporal smoothing, and $\eta=1.0$ is the curiosity scale. By focusing on geometric task structure, POE substantially reduces spurious exploration caused by lighting changes and camera motion. The base progress term $r^{\text{base}}_t$ is forwarded to Long-Horizon Memory for memory-conditioned gating before contributing to the combined reward.

\subsection{Long-Horizon Memory}\label{sec:mcis}

\noindent\textbf{Memory Mechanism: Beyond Adjacent-Averaging Consolidation.}
Long-horizon tasks challenge memory management. Memory-augmented architectures range from Neural Map~\cite{parisotto2018neural} to recent VLA-oriented MemoryVLA~\cite{memoryvla2025}, yet many still compress sequences via averaging or truncation, which blurs stage-specific details and weakens temporal identifiability. In high-dimensional embodied settings, accumulated long-tail noise can amplify stage hallucination and hinder credit assignment in multi-step tasks. We instead treat \emph{memory as context}: at each step we select Top-$K$ history items as independent context tokens, fuse them with the current latent via a learned gate, and perform lightweight write-back with utility-guided eviction (usage, recency, redundancy) rather than adjacent averaging. This retains phase-specific information while keeping the store compact.

\noindent\textbf{Context Selection via Attention.}
We extract a latent representation $x_t \in \mathbb{R}^d$ from the OpenVLA-OFT backbone at each step (by pooling over the LLM's hidden states), and we maintain a memory store $\mathcal{M} = \{m_i \in \mathbb{R}^d\}_{i=1}^{L}$ with timestamps $\{\tau_i\}$, where $L$ is the store capacity. To select relevant history, we define query, key, and value projections:
\begin{equation}
\begin{aligned}
q_t &= W_q x_t, \\
k_i &= W_k\!\big(m_i + \text{PE}(\tau_i)\big), \\
v_i &= W_v m_i,
\end{aligned}
\end{equation}
where $W_q, W_k, W_v \in \mathbb{R}^{d \times d}$ are learnable linear maps, and $\text{PE}(\cdot)$ is a sinusoidal temporal positional encoding. Attention scores and the selected index set are computed as:
\begin{equation}
\label{eq:lhm-attn}
\begin{aligned}
a_i &= \frac{\exp(\langle q_t, k_i \rangle / \sqrt{d})}{\sum_{j=1}^{L} \exp(\langle q_t, k_j \rangle / \sqrt{d})}, \\
\mathcal{S}_t &= \text{TopK}\big(\{a_i\}_{i=1}^{L}, K\big),
\end{aligned}
\end{equation}
where $K$ selects the most relevant history items. We prepend $\{v_i \mid i \in \mathcal{S}_t\}$ to the current sequence as independent context tokens, replacing static adjacent-averaging consolidation used in prior work.

\noindent\textbf{Gated Fusion and Reward Modulation.}
Let $\hat{h}_t = \sum_{i \in \mathcal{S}_t} a_i v_i$ be the weighted context embedding. We fuse it with the current latent $x_t$ via a learned gate:
\begin{equation}
\begin{aligned}
g^{\text{mem}}_t &= \sigma\!\big(w_g^{\top} [\hat{h}_t; x_t]\big), \\
\tilde{x}_t &= (1 - g^{\text{mem}}_t) x_t + g^{\text{mem}}_t \hat{h}_t.
\end{aligned}
\end{equation}
where $w_g \in \mathbb{R}^{2d}$ is a learnable weight vector and $[\cdot;\cdot]$ denotes concatenation. The gate $g^{\text{mem}}_t$ modulates how much history context influences the current representation. The memory gate also modulates the base progress reward $r^{\text{base}}_t$:
\begin{equation}
r^{\text{prog}}_t = g^{\text{mem}}_t \cdot r^{\text{base}}_t.
\end{equation}
suppressing spurious progress signals when history context indicates unstable manipulation patterns (e.g., repeated failures, oscillations). We block gradients through $g^{\text{mem}}_{t-1}$ to prevent reward hacking. Memory update writes back the fused representation $\tilde{x}_t$ only to the selected items $\mathcal{S}_t$, and eviction removes the least useful entry based on a utility score combining usage frequency, recency, and redundancy with existing items.

\noindent\textbf{POE Integration and Effect.}
In contrast to Titans-MAC~\cite{behrouz2024titans} and MemoryVLA~\cite{memoryvla2025}, Long-Horizon Memory uses selective context with gated fusion for reward-level modulation. The core mechanism operates on backbone latents $x_t$ and history items $m_i$ via the attention and gating mechanisms above. When combined with POE (full model, Table~\ref{tab:hallucination} row 5 vs.\ row 4 standalone), POE's geometric pose tracking provides complementary grounding that enhances memory's ability to distinguish task-relevant manipulation contexts from spurious visual changes, further improving robustness. This mechanism complements SAR temporal smoothing and aligns with the training objective $\mathcal{L}$~\eqref{eq:total_loss}. Empirically, Long-Horizon Memory in standalone mode contributes +2.4 points of success and a 3.9-point hallucination reduction, with POE integration adding an additional +3.1 points, highlighting the value of both selective context and geometric grounding.

\subsection{Training Objective}

The full training objective combines policy gradient loss, value losses, and world model losses:
\begin{equation}
\label{eq:total_loss}
\mathcal{L} = \mathcal{L}_{\text{PPO}}(\tilde{r}) + \lambda_F \mathcal{L}_F + \lambda_I \mathcal{L}_I - \lambda_{\text{ent}} H(\pi).
\end{equation}
where $\mathcal{L}_F$ and $\mathcal{L}_I$ are the MSE losses for the forward and inverse world models, $H(\pi)$ is policy entropy, and $\lambda_F, \lambda_I, \lambda_{\text{ent}}$ weight the auxiliary objectives (see Table~\ref{tab:hyperparams} for values). We use PPO (Proximal Policy Optimization)~\cite{schulman2017proximal} with dual critics $V_e, V_i$ for extrinsic/intrinsic returns, fusing advantages as $A_t = (1-\rho) A^e_t + \rho A^i_t$ using GAE (Generalized Advantage Estimation)~\cite{schulman2016high}. World models and the memory-conditioned GRU are trained via separate optimizers with stop-gradient to prevent reward exploitation.

\section{Discoverse-L Benchmark}\label{sec:dataset}

\noindent\textbf{Benchmark Overview.}
Discoverse-L builds on the DISCOVERSE simulator~\cite{discoverse2025arxiv} with the AIRBOT-Play platform~\cite{airbotplay}, offering 3 multi-stage manipulation tasks with varying difficulty: (1) \textit{Block Bridge} (74 stages): place two bars to form a bridge structure, then fill with multiple blocks, the most complex task requiring precise multi-object coordination; (2) \textit{Stack} (18 stages): stack three colored blocks in sequence, demanding accurate alignment; (3) \textit{Jujube-Cup} (19 stages): place a jujube fruit into a cup and move the cup onto a plate, a relatively simpler setup with fewer contact-critical stages. Stage counts reflect our Gemini~\cite{gemini2024api}-prompted dictionaries; the DISCOVERSE simulator provides fine-grained ground-truth annotations (e.g., 79 for Bridge) used only for validation of stage-discovery coverage. The 74-stage Bridge dictionary merges five simulator micro-adjustments into semantically equivalent steps (93.7\% recall; Appendix~\ref{app:templates}), so reward supervision remains faithful while prompts stay compact. We generate 50 simulator rollouts per task with randomized initial conditions (Supp. Mat. Appendix~\ref{app:dataset-details} provides task-aligned normalization and evaluation protocol details).

\section{Experiments}\label{sec:results}

\subsection{Experimental Design and Evaluation}

For comparability across methods on Discoverse-L, we recommend initializing policies with the provided task-aligned normalization and optionally bootstrapping from the 50 demonstrations via supervised learning before engaging in interactive PPO training. Evaluation typically uses 50 interactive episodes per task with held-out seeds, identical camera setups, and fixed episode budgets.

\begin{figure*}[t]
    \centering
    \includegraphics[width=\linewidth]{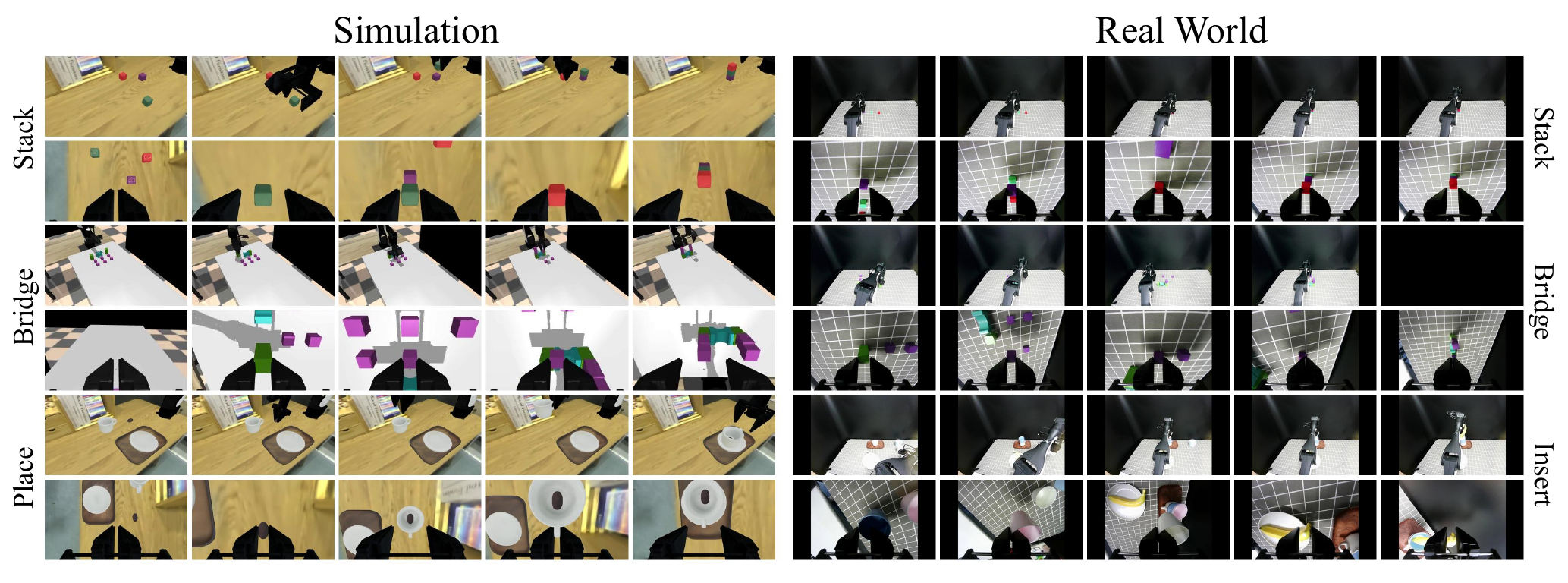} %
\caption{\textbf{Simulation and real-world rollouts.}
Columns depict temporal progression, rows pair the same task family across domains. Left: dual-camera Discoverse-L rollouts for Stack, Bridge, Jujube-Cup; right: AIRBOT-Play deployments for Stack, Bridge, Insert. The alignment highlights consistent gripper–object interactions and viewpoints after Sim2Real transfer.}
    \vspace{-0.4cm}
    \label{fig:simulation}
\end{figure*}

\noindent\textbf{Implementation Details.}\label{sec:implementation}
We build upon the OpenVLA-OFT architecture~\cite{kim2025openvlaoft}, which extends the original OpenVLA~\cite{kim2024openvla} with optimized fine-tuning techniques including parallel action decoding, action chunking, and L1 regression loss. The policy backbone follows OpenVLA-OFT: multi-view ViT towers (SigLIP, ICCV 2023~\cite{zhai2023sigmoid} + DINOv2, arXiv 2023~\cite{oquab2023dinov2}) and a language model (Llama-2 7B~\cite{touvron2023llama}) with FiLM modulation and chunked parallel action decoding. For stage evaluation we use a separate frozen CLIP encoder (ViT-B/16~\cite{radford2021clip}, 224$\times$224 input) to compute image–text similarity for SAR; separating the evaluator prevents reward overfitting. The HR computation uses the same frozen CLIP with threshold $\theta=0.7$, selected via sensitivity analysis (Supp. Mat. Fig.~\ref{fig:threshold-sensitivity}) to balance false-positive and false-negative rates. World models for pose-based exploration are lightweight MLPs. Long-Horizon Memory uses a single-layer GRU with a sigmoid readout to modulate the progress reward. Training uses 8 parallel DISCOVERSE environments with PPO for 2M timesteps per seed across 3 random seeds. Hardware: 4$\times$NVIDIA H20 GPUs (96 GB VRAM each); wall-clock training time is approximately 24 hours per seed. Complete hyperparameters are provided in the Supp. Mat. %

Stage dictionaries are generated via the video-driven discovery workflow described above: for the three Discoverse-L tasks, we use ten simulator demonstrations per task; for the unseen real-world assembly task, we run the same pipeline on its 50 teleoperated demonstrations. The resulting semantics supervise both stage gating and SAR triplet selection across domains. We also load the task-aligned normalization statistics from Discoverse-L and inject them at policy initialization (matching the active task) to ensure feature magnitudes remain consistent between training and evaluation. Following prior practice, many implementations optionally apply a short behavior cloning warm-start before interactive RL fine-tuning. All simulation results report mean over three independent random seeds.

\begin{table}[!tb]
\centering
\small
\caption{Discoverse-L benchmark (3 tasks; mean of 3 seeds, 50 evaluation episodes). All models use task-aligned normalization; EvoVLA adds SAR, POE, and Long-Horizon Memory to OpenVLA-OFT. The last row highlights improvements over OpenVLA-OFT.}\label{tab:main-results}
\begin{tabular}{lcccc}
\toprule
\textbf{Model} & \textbf{Bridge} & \textbf{Jujube-Cup} & \textbf{Stack} & \textbf{Avg.} \\
\midrule
Octo~\cite{octo2024} & 24.8 & 33.7 & 29.1 & 29.2 \\
OpenVLA~\cite{kim2024openvla} & 32.6 & 42.0 & 37.5 & 37.4 \\
$\pi_0$~\cite{black2024pi0,openpi2024} & 41.1 & 50.5 & 46.2 & 45.9 \\
$\pi_0$-FAST~\cite{openpi2024,fast2025} & 47.4 & 56.9 & 52.8 & 52.4 \\
OpenVLA-OFT~\cite{kim2025openvlaoft} & 54.1 & 63.5 & 59.4 & 59.0 \\
\midrule
\textbf{EvoVLA (OURS)} & \textbf{65.3} & \textbf{72.6} & \textbf{69.7} & \textbf{69.2} \\
\,\,$\Delta$ vs.\ OFT & \textcolor{green}{+11.2} & \textcolor{green}{+9.1} & \textcolor{green}{+10.3} & \textcolor{green}{+10.2} \\
\bottomrule
\end{tabular}
    \vspace{-0.4cm}
\end{table}

\noindent\textbf{Evaluation Metrics.}\label{sec:metrics}
\noindent\emph{Success Rate (SR):} Percentage of episodes completing the task within 400 steps. We cap episodes at 400 to accommodate the most complex task (Bridge, 74 stages) while preventing indefinite rollouts; successful runs typically finish in 50–100 steps for simpler tasks (Stack, Jujube-Cup) and 150–250 steps for Bridge.

\noindent\emph{Sample Efficiency:} Number of environment timesteps required to reach a 50\% success rate.

\noindent\emph{Hallucination Rate (HR):} The fraction of high VLM scores corresponding to failed actual completion. Formally, let $u_k(t) \in [0,1]$ be the VLM score for stage $k$ at time $t$, and $c_k(t) \in \{0,1\}$ be the ground-truth completion indicator. Hallucination rate (HR) is:
\begin{equation}
\text{HR} = \frac{\mathbb{E}\big[\ind{u_k(t) > \theta \land c_k(t) = 0}\big]}{\mathbb{E}\big[\ind{u_k(t) > \theta}\big]}.
\end{equation}
where $u_k(t) > \theta$ (we use $\theta=0.7$) indicates ``high VLM score'' and $c_k(t) = 0$ indicates failed completion, verified using task success signals from the DISCOVERSE simulator. The expectation averages over evaluation episodes, timesteps, and active stages. We emphasize that $c_k(t)$ is obtained solely at evaluation time from DISCOVERSE stage-event triggers to compute the metric; it is never used for training or model selection.

\noindent\textbf{Real-World Deployment.}
Beyond simulation, we validate EvoVLA on a physical AIRBOT-Play robot (details in Section~\ref{sec:real-world}).

\subsection{Simulation Results}

\noindent\textbf{Baselines.}
We benchmark against state-of-the-art VLA backbones trained on Discoverse-L: Octo~\cite{octo2024}, OpenVLA~\cite{kim2024openvla}, $\pi_0$ and $\pi_0$-FAST~\cite{openpi2024,fast2025}, and OpenVLA-OFT~\cite{kim2025openvlaoft}. EvoVLA is obtained by augmenting the OpenVLA-OFT architecture with our SAR module, POE module, Long-Horizon Memory module, and task-aligned normalization to address stage hallucination. %

\noindent\textbf{Main Simulation Results.}
Table~\ref{tab:main-results} summarizes quantitative comparisons against the baselines. Starting from Octo (29.2\%) and OpenVLA (37.4\%), progressively stronger models achieve 45.9\% ($\pi_0$), 52.4\% ($\pi_0$-FAST), and 59.0\% (OpenVLA-OFT) through richer pretraining and architectural improvements. EvoVLA lifts average success to 69.2\%, improving upon the strongest baseline by +10.2 absolute points. Per-task analysis shows consistent gains: Bridge (+11.2 points, 65.3\% vs.\ 54.1\%), Jujube-Cup (+9.1 points, 72.6\% vs.\ 63.5\%), and Stack (+10.3 points, 69.7\% vs.\ 59.4\%), demonstrating that our approach generalizes across tasks with varying complexity (18--74 stages) while using the same backbone architecture. Figure~\ref{fig:simulation} complements these numbers by juxtaposing dual-camera rollouts in simulation and on the AIRBOT-Play robot, highlighting that EvoVLA maintains consistent gripper-object interactions across domains. Full-resolution sequences for simulation-only and real-world-only views are provided in Supp. Mat. Figures~\ref{fig:sim-rollouts-supp} and~\ref{fig:real-rollouts-supp}.

\noindent\textbf{Sample Efficiency.}
EvoVLA reaches a 50\% average success rate in approximately $6\times 10^5$ environment steps (averaged over 3 seeds), whereas OpenVLA-OFT requires approximately $9\times 10^5$ steps, demonstrating 1.5$\times$ better sample efficiency. This acceleration stems from dense intrinsic feedback: SAR provides stage-level guidance at every timestep, POE grounds exploration in task-relevant geometric structures, and Long-Horizon Memory stabilizes learning signals.

\noindent\textbf{Hallucination Reduction.}
Simulation rollouts report a 14.8\% Hallucination Rate versus 38.5\% for OpenVLA-OFT, i.e., a 23.7-point drop (Table~\ref{tab:hallucination}). The precise contribution of each module—hard negatives, temporal smoothing, Long-Horizon Memory, and POE—is analyzed in Section~\ref{sec:ablations}, so we defer further discussion there.

\subsection{Real-World Evaluation}\label{sec:real-world}

We deploy EvoVLA on a physical AIRBOT-Play robot (additional rollout snapshots shown in Figure~\ref{fig:simulation}) to assess its real-world performance.

\begin{figure*}[!t]
    \centering
    \includegraphics[width=\linewidth]{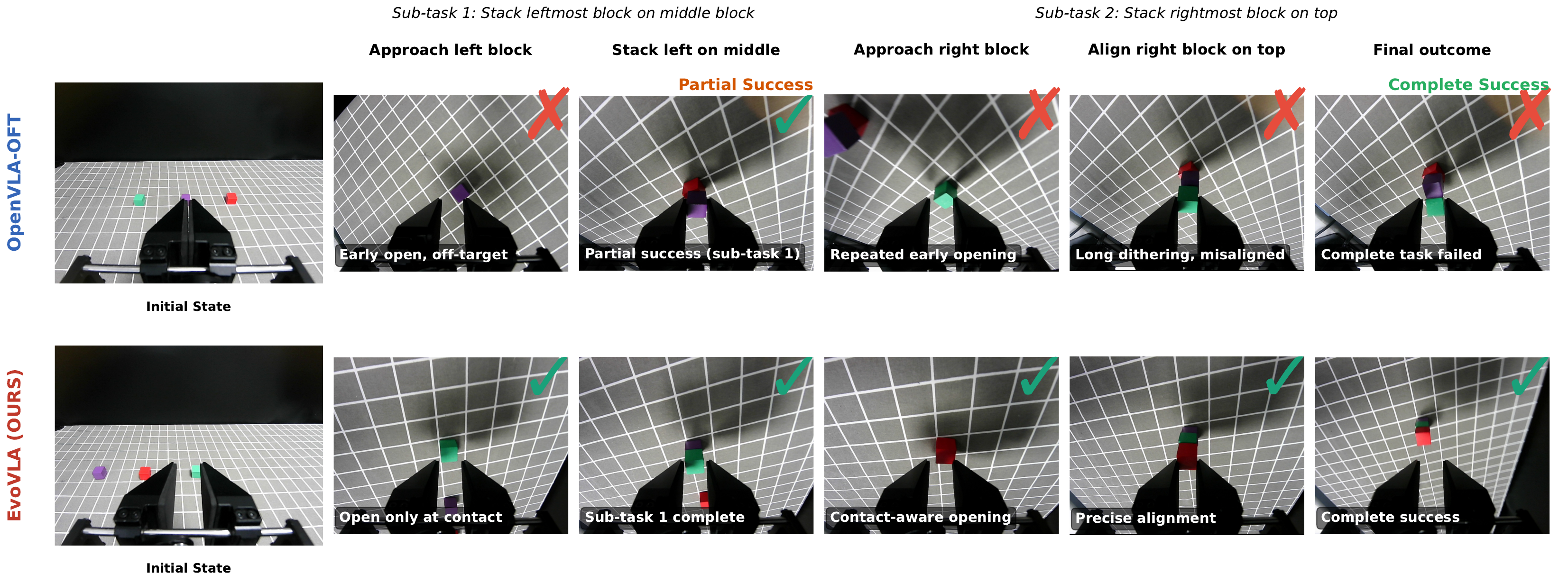}
\caption{\textbf{Stack task qualitative comparison.} Columns (1--5) cover sub-task 1 (left block onto middle) then sub-task 2 (right block on top). \textit{Top:} OpenVLA-OFT (cross markers) opens before contact, dithers, misaligns, and drops the block. \textit{Bottom:} EvoVLA (check marks) delays opening until contact, aligns within a few corrections, and leaves a stable stack, matching the hallucination reductions in Section~\ref{sec:real-world}.}
    \label{fig:stack-qualitative}
\end{figure*}

We benchmark against two baselines: $\pi_0$-FAST~\cite{openpi2024,fast2025} and OpenVLA-OFT~\cite{kim2025openvlaoft}, both trained with the same real-world setup. Evaluation comprises two parts: Sim2Real transfer of simulation-trained policies to the three Discoverse-L tasks, and training a new policy from scratch on an unseen assembly task.

\noindent\textbf{Sim2Real Transfer (Three Tasks).}
For the three Discoverse-L tasks (Block Bridge, Jujube-Cup, Stack), we directly deploy the simulation-trained policies to the physical robot without any real-world retraining. Each policy receives only a language prompt describing the task (e.g., ``Build block bridge: place two bars, fill with blocks''). Third-person and wrist-mounted camera recordings confirm successful task completion across all three scenarios, demonstrating effective Sim2Real transfer enabled by the video-driven stage semantics and task-aligned normalization learned in simulation.

\noindent\textbf{Unseen Task: Cup Stacking with Insertion.}
We further evaluate a novel assembly task that stacks four cups and inserts a banana-shaped object into the final stack. Fifty teleoperated demonstrations provide stage semantics via Gemini~2.5~Pro and refreshed normalization statistics. The policy is warm-started with BC, then fine-tuned on-robot with PPO~\cite{schulman2017proximal} for roughly 5k steps under safety monitoring. During fine-tuning, SAR queries CLIP~\cite{radford2021clip} against the wrist view, POE tracks AprilTag-based~\cite{wang2011apriltag} relative poses, and Long-Horizon Memory gates the progress reward, with third-person and wrist-camera videos documenting successful execution.

\begin{figure}[!t]
    \centering
    \scalebox{1.15}{\includegraphics[width=0.9\linewidth]{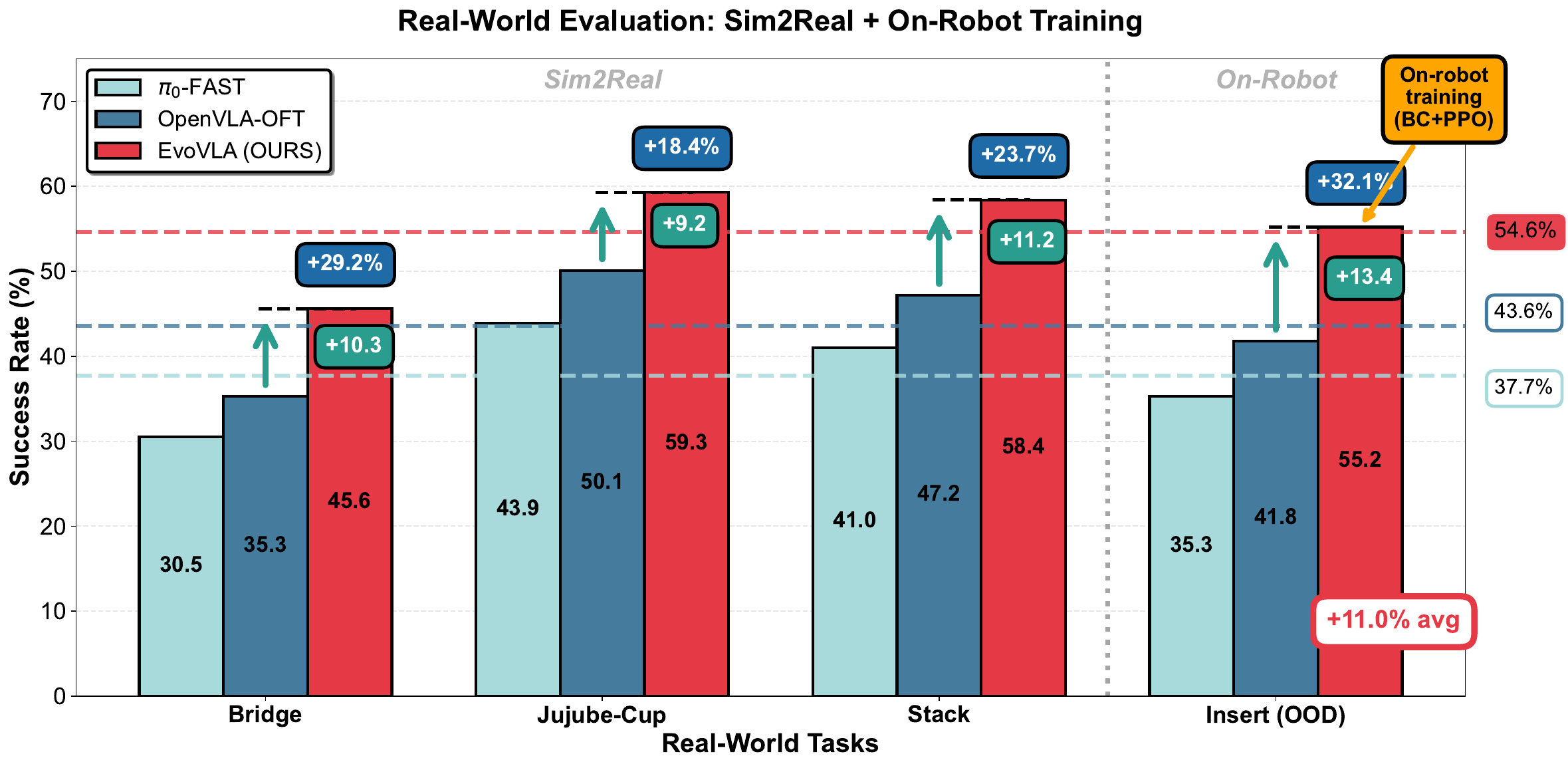}}
    \caption{\textbf{Real-world evaluation.} Green arrows: absolute gains over OpenVLA-OFT (+9.2--+13.4 points). Blue labels: relative gains (+18.4\%--+32.1\%). Horizontal dashed lines mark model averages (37.7\%, 43.6\%, 54.6\%); the vertical dashed line splits Sim2Real transfers (left) from the on-robot Insert task (right).}
        \vspace{-0.4cm}
        \label{fig:real-world}
\end{figure}

\noindent\textbf{Main Real-World Results.}
Figure~\ref{fig:real-world} and Table~\ref{tab:real-world} show that EvoVLA attains 54.6\% average success across the three Sim2Real tasks plus the on-robot insertion task, surpassing OpenVLA-OFT by +11.0 points overall and by +9.2--+13.4 points on each task. The unseen insertion task sees the largest gain (55.2\% vs.\ 41.8\%). Third-person and wrist-mounted recordings confirm successful executions for all tasks, highlighting consistent Sim2Real behavior.

\noindent\textbf{Real-World Qualitative Case Study.}
Figure~\ref{fig:stack-qualitative} provides a qualitative case study on the Stack task. OpenVLA-OFT exhibits premature gripper opening, prolonged dithering, and misalignment (top-row columns marked with cross symbols), whereas EvoVLA maintains contact-aware control and precise placement (bottom-row check marks), yielding a stable three-block tower. These observations align with the quantitative Stack gains in Table~\ref{tab:hallucination} and reinforce that stage-aligned shaping and pose-grounded exploration mitigate stage hallucination on real robots.

\subsection{Ablation Studies}\label{sec:ablations}

\begin{table}[!htb]
\centering
\small
\caption{\textbf{Component ablation.} Progressive addition (averaged across 3 tasks, 3 seeds). Starting from OpenVLA-OFT baseline, we cumulatively add: hard negatives, temporal smoothing, Long-Horizon Memory, and POE.}\label{tab:hallucination}
\begin{tabular}{lcc}
\toprule
\textbf{Method} & \textbf{SR (\%)} & \textbf{HR (\%)} \\
\midrule
OpenVLA-OFT (Baseline) & 59.0 & 38.5 \\
+ Hard Negatives ($T^{h-}$) & 61.8 & 31.2 \\
+ Temporal Smoothing & 63.7 & 23.4 \\
+ Long-Horizon Memory & 66.1 & 19.5 \\
+ Pose-based Exploration & \textbf{69.2} & \textbf{14.8} \\
\bottomrule
\end{tabular}
        \vspace{-0.4cm}
\end{table}

Table~\ref{tab:hallucination} quantifies the cumulative gains: hard negatives add +2.8 SR / -7.3 HR by forcing discrimination of near-miss states; temporal smoothing contributes +1.9 / -7.8 by filtering score spikes; Long-Horizon Memory supplies +2.4 / -3.9 via selective context retention; POE adds +3.1 / -4.7 by grounding curiosity in task geometry. Together they deliver +10.2 points SR and -23.7 points HR over OpenVLA-OFT, and removing any single component forfeits at least 2.4 success points. Figure~\ref{fig:stack-qualitative} visualizes these improvements on the Stack task. 

\begin{figure}[H]
    \centering
    \includegraphics[width=\linewidth]{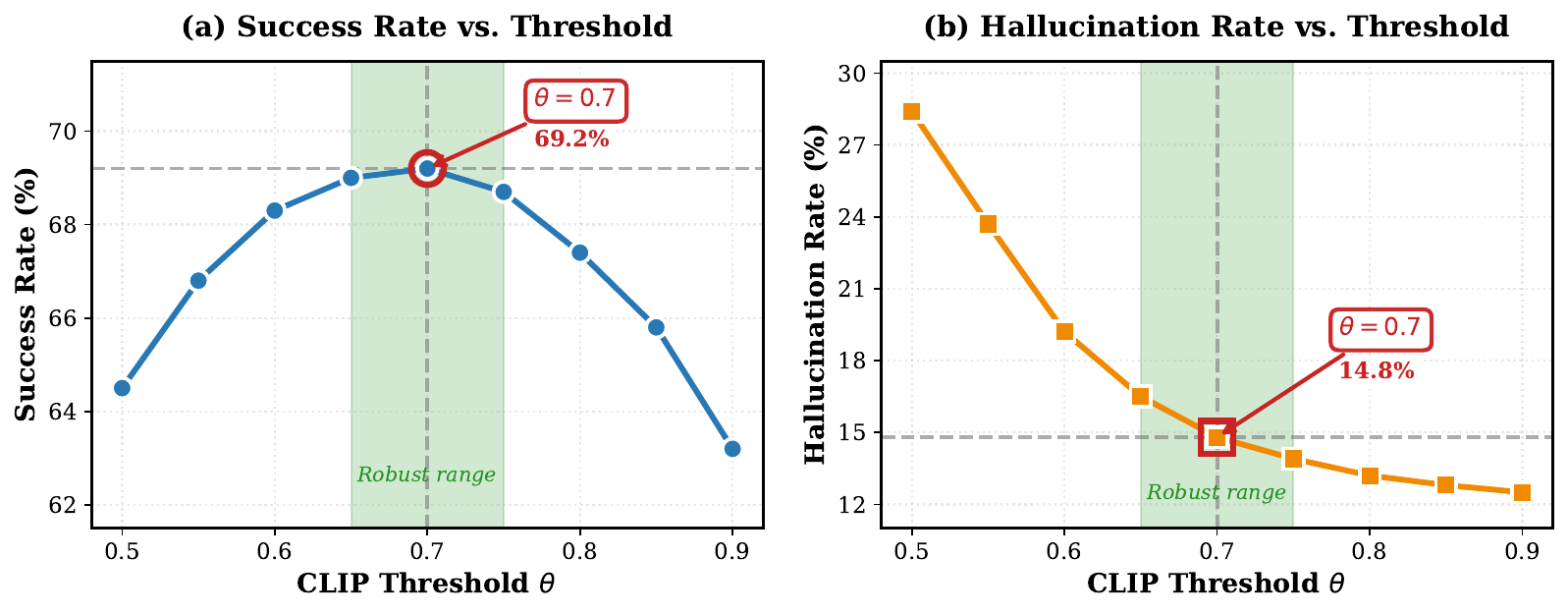}
    \caption{CLIP threshold sensitivity for EvoVLA (full model) averaged across three Discoverse-L tasks and three seeds. (a) Success Rate remains stable within $\pm$0.5 points around $\theta=0.7$ (68.7--69.2\%). (b) Hallucination Rate decreases monotonically with higher thresholds; $\theta=0.7$ balances false-positive and false-negative rates while maintaining peak success. Results demonstrate that our choice is robust to threshold variations within [0.65, 0.75].}\label{fig:threshold-sensitivity}
\end{figure}

\begin{figure}[H]
    \centering
    \includegraphics[width=\linewidth]{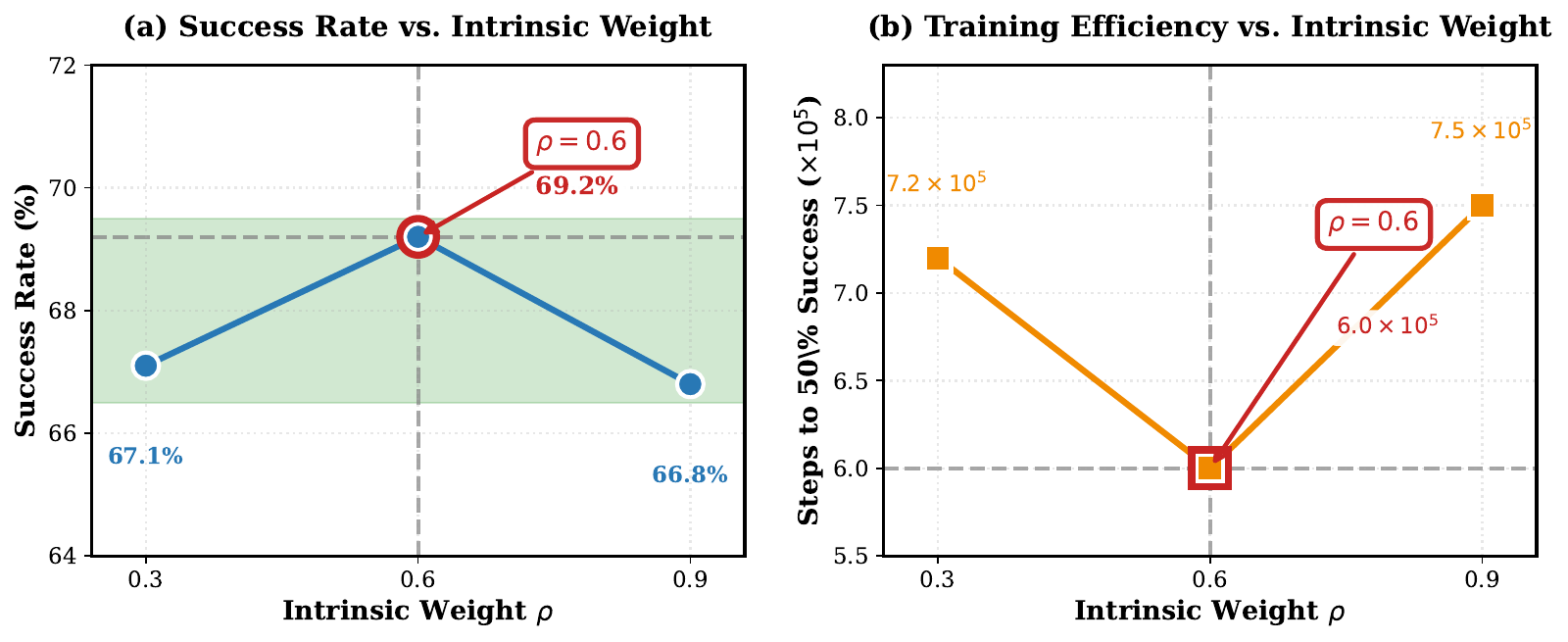}
    \caption{\textbf{Intrinsic weight $\rho$ sensitivity.} Performance with $\rho \in \{0.3, 0.6, 0.9\}$ (averaged across 3 tasks, 3 seeds). (a) Success Rate: stable within $\pm 2.5\%$, peaking at $\rho=0.6$ (69.2\%). (b) Training efficiency: $\rho=0.6$ achieves fastest convergence to 50\% success. Results confirm robustness to intrinsic weight selection.}\label{fig:rho-sensitivity}
\end{figure}

Hyperparameter sensitivity analyzes (Supp. Mat. Figures~\ref{fig:threshold-sensitivity}, \ref{fig:rho-sensitivity}) demonstrate robustness: the CLIP threshold $\theta$ remains effective within [0.65, 0.75], and intrinsic weight $\rho$ shows stable performance ($\pm 2.4$ points) across [0.3, 0.9], confirming that our choices of $\theta=0.7$ and $\rho=0.6$ are well-justified.

\section{Conclusion}\label{sec:conclusion}

We identified and addressed stage hallucination, where policies fool VLM evaluators without task completion. Our EvoVLA framework introduces stage-aligned rewards with triplet contrastive learning and temporal smoothing, long-horizon memory that stabilizes intrinsic shaping over extended interactions, and pose-based exploration, which grounds curiosity in the relative object-gripper pose space. These components work synergistically to prevent visual shortcuts and substantially reduce spurious exploration. Comprehensive ablation studies demonstrate that each component contributes significantly to performance. Experiments demonstrate a 69.2\% average success rate (+10.2 points over the strongest baseline), 1.5$\times$ improved sample efficiency (reaching 50\% success with fewer environment steps), and a Hallucination Rate reduction to 14.8\% across the three Discoverse-L tasks.

\clearpage
\setcounter{page}{1}
\maketitlesupplementary%

\section{Limitations and Discussion}\label{sec:discussion}

\paragraph{Limitations}
\begin{itemize}[leftmargin=1.5em,itemsep=0.05em]
  \item Sim2Real gap: Quantitative simulation results are complemented by real-robot demonstrations; systematic real-world evaluation remains future work.
  \item Dependency on 6D poses: Requires accurate object tracking (simulator ground-truth or AprilTag markers). Future work will integrate learned 6D pose estimation~\cite{wang2021gdr}.
  \item Video-driven stage discovery coverage: Automatic stage dictionaries reduce manual verification, yet still rely on demonstration diversity and prompt quality; rare corner cases may require additional data curation.
  \item Compute cost: Training requires 4$\times$H20 GPUs$\times$24h per seed (72h total for 3 seeds). Optimizing sample efficiency remains important for scalability.
\end{itemize}

\paragraph{Threats to Validity}
Results may vary under dataset or simulator changes (e.g., asset versions), implementation choices for baselines (pre-processing and augmentations), or unseen distribution shifts. We mitigate these threats via strict budget matching (2M steps across methods), identical backbones and input resolutions, and averaging over 3 independent random seeds.

\section{Reproducibility Statement}\label{sec:reproducibility}

We will release EvoVLA training and evaluation code, stage dictionaries, and RLDS-formatted trajectories upon publication. Experiments were conducted on 4$\times$NVIDIA H20 GPUs with deterministically seeded runs. Table~\ref{tab:hyperparams} provides complete hyperparameter settings for all components.

\section{Real-World Detailed Results}

\begin{table}[H]
\centering
\footnotesize
\caption{Real-world evaluation on four tasks. Each model--task pair uses at least 20 valid physical trials (episodes without safety aborts); some configurations include additional valid trials, yielding one-decimal precision. The three Discoverse-L tasks deploy simulation-trained policies, whereas the insertion task is trained on-robot (BC warm-start + PPO).}\label{tab:real-world}
\begin{tabular}{l@{\hspace{0.3em}}c@{\hspace{0.3em}}c@{\hspace{0.3em}}c@{\hspace{0.3em}}c@{\hspace{0.3em}}|@{\hspace{0.3em}}c} %
\toprule
\textbf{Model} & \textbf{Bridge} & \textbf{Jujube-Cup} & \textbf{Stack} & \textbf{Insert (OOD)} & \textbf{Avg.} \\
\midrule
$\pi_0$-FAST~\cite{openpi2024,fast2025} & 30.5 & 43.9 & 41.0 & 35.3 & 37.7 \\
OpenVLA-OFT~\cite{kim2025openvlaoft} & 35.3 & 50.1 & 47.2 & 41.8 & 43.6 \\
\midrule
\textbf{EvoVLA (OURS)} & \textbf{45.6} & \textbf{59.3} & \textbf{58.4} & \textbf{55.2} & \textbf{54.6} \\
\bottomrule
\end{tabular}
\end{table}

\section{Additional Rollout Visualizations}

\begin{figure*}[t]
    \centering
    \includegraphics[width=\linewidth]{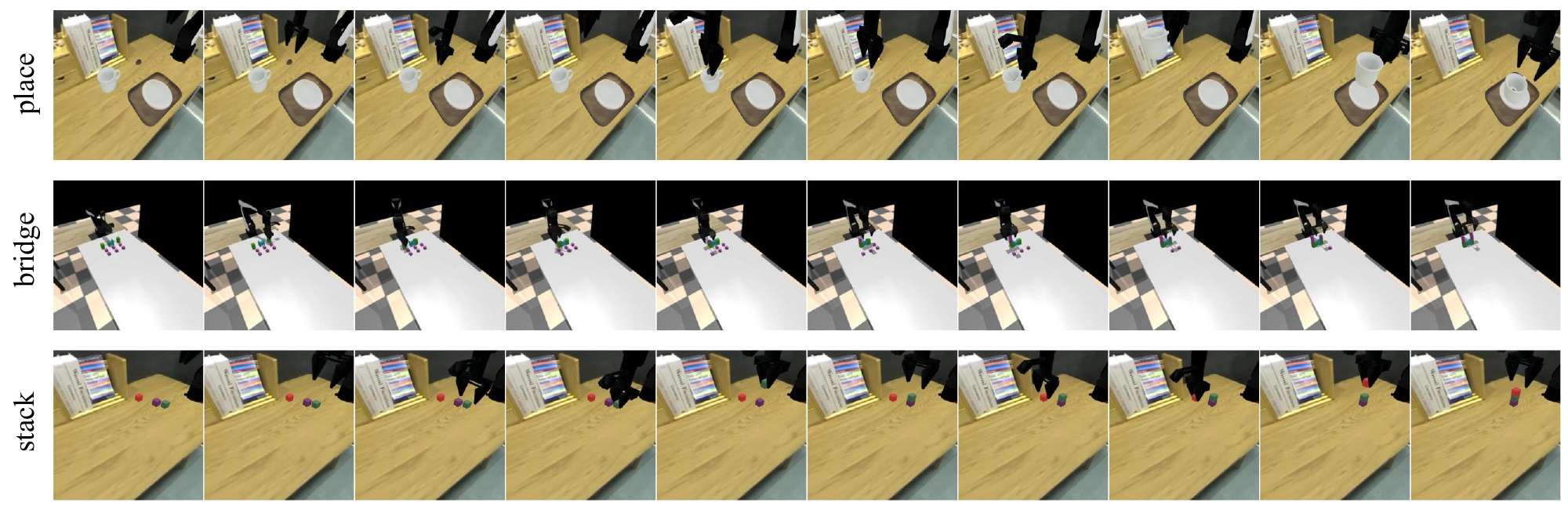}
    \caption{Supplementary Discoverse-L rollouts. Each row shows a full episode for place, bridge, and stack with dual third-person/wrist views; columns denote temporal progression.}
    \label{fig:sim-rollouts-supp}
\end{figure*}

\begin{figure*}[t]
    \centering
    \includegraphics[width=\linewidth]{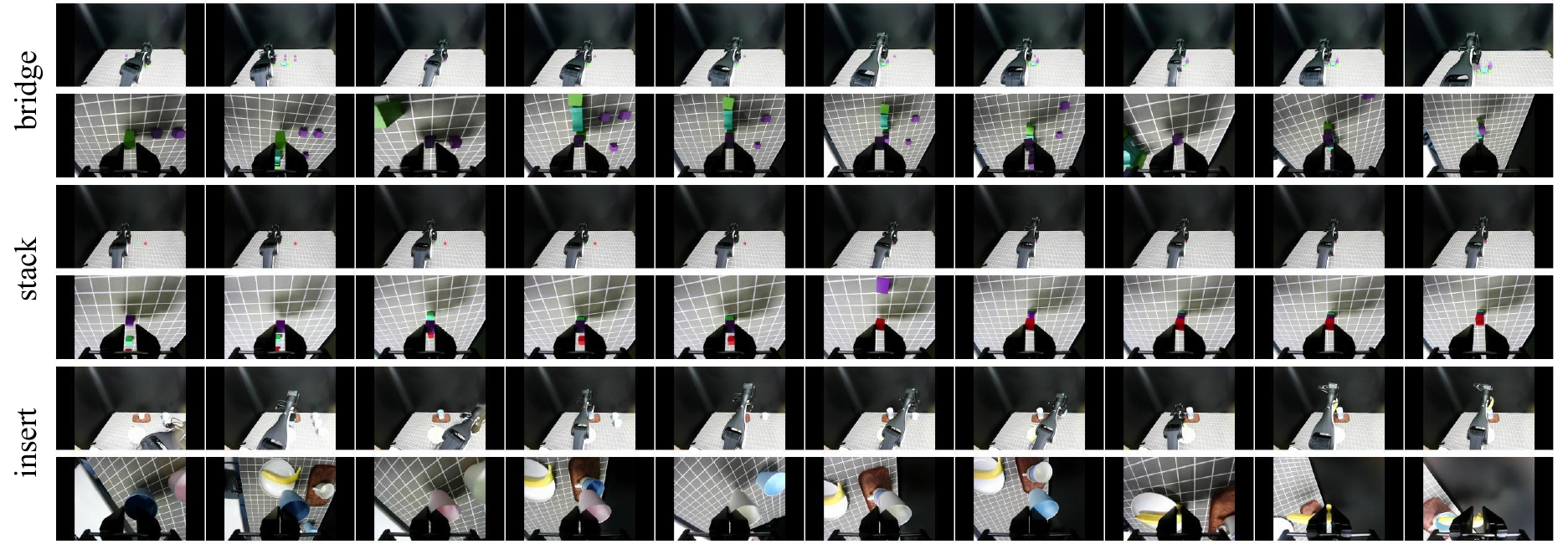}
    \caption{Supplementary AIRBOT-Play rollouts. Each row shows stack, bridge, and insert executions with synchronized third-person and wrist cameras in the real world.}
    \label{fig:real-rollouts-supp}
\end{figure*}

This section expands the qualitative evidence from Figure~\ref{fig:simulation} by presenting the full rollout sequences. Figure~\ref{fig:sim-rollouts-supp} provides high-resolution Discoverse-L execution traces for the place, bridge, and stack tasks, illustrating how EvoVLA maintains stable gripper-object contact throughout each stage. Figure~\ref{fig:real-rollouts-supp} mirrors this analysis on the AIRBOT-Play platform for stack, bridge, and insert executions, highlighting consistent Sim2Real behavior across synchronized third-person and wrist-mounted views.

\section{Problem Formulation}

\paragraph{POMDP Setting}
We model long-horizon robotic manipulation as a partially observable Markov decision process (POMDP) defined by the tuple $(\mathcal{S}, \mathcal{A}, \mathcal{O}, T, R, \Omega, \gamma)$, where $\mathcal{S}$ is the latent state space (object poses and robot configuration), $\mathcal{A}$ the continuous action space (6-DoF end-effector twist deltas plus a 1-D gripper command; total 7-D), $\mathcal{O}$ the observation space (multi-view RGB + proprioception), $T\colon \mathcal{S} \times \mathcal{A} \rightarrow \mathcal{S}$ the transition function, $R\colon \mathcal{S} \times \mathcal{A} \rightarrow \mathbb{R}$ the extrinsic reward (sparse task success), $\Omega\colon \mathcal{S} \rightarrow \mathcal{O}$ the observation function, and $\gamma \in [0,1)$ the discount factor.%

\paragraph{Task Specification and Stage Decomposition}
Each task is specified by a natural-language instruction $\ell$ (e.g., ``Build block bridge: place two bars, fill with blocks'') and decomposes into $K$ sequential stages indexed by $k \in \{1, \ldots, K\}$. We obtain this decomposition via video-driven discovery: given $N$ demonstration rollouts for task $\ell$, we apply Gemini~2.5~Pro~\cite{gemini2024api} to align them and extract a stage dictionary
\begin{equation}
\mathcal{D} = \big\{(T^+_k,\, T^-_k,\, T^{h-}_k)\big\}_{k=1}^{K}.%
\end{equation}
where $T^+_k$ is a text predicate describing successful completion of stage $k$ (e.g., ``gripper contacts target bar; fingers closed; stable grasp''), $T^-_k$ is a mutually exclusive failure predicate (e.g., ``gripper far from all objects; fingers open at init''), and $T^{h-}_k$ is a counterfactual near-miss predicate that differs by a single spatial or contact constraint (e.g., ``gripper near target bar but no contact; fingers open''). See Appendix~\ref{app:templates} for complete examples from the Bridge task (Gemini produced 74 stages out of 79 simulator ground-truth stages), Stack task (18 stages), and Jujube-Cup task (19 stages).

During policy execution, we maintain a stage tracker that consumes the observation history $\mathcal{H}_t = \{o_0, \ldots, o_t\}$ and outputs the current stage index $\kappa_t$. The tracker advances $\kappa_t \rightarrow \kappa_t+1$ when a sliding window of $m=8$ recent stage rewards (computed via image-text similarity between $o_t$ and $T^+_{\kappa_t}$) exceeds a threshold, as detailed in Appendix~\ref{app:implementation}. Crucially, no ground-truth stage labels from the simulator are consumed during training or evaluation; stage progression relies solely on the learned VLM scores and the stage dictionary semantics. In practice, the stage dictionary $\mathcal{D}$ supervises SAR during simulation training and is not re-estimated for Sim2Real deployment on the same three real-world tasks; the learned policy implicitly internalizes these semantics. For the unseen real-world task trained on-robot, we generate its own dictionary from 50 teleoperated demonstrations.

\noindent\textbf{Policy objective.}
The agent learns a policy $\pi_\theta\colon \mathcal{O} \times \{1, \ldots, K\} \rightarrow \mathcal{A}$ that conditions on the current observation and stage index. We maximize the expected discounted return
\begin{equation}
J(\theta) = \mathbb{E}_{\xi \sim \pi_\theta}\Bigg[\sum_{t=0}^{H} \gamma^{t} \, \tilde{r}_t\Bigg].
\end{equation}
where $\xi = (o_0, a_0, \tilde{r}_0, \ldots)$ is a trajectory, $H$ is the episode horizon (400 steps in our DISCOVERSE experiments), $\gamma=0.995$ is the discount factor (standard for long-horizon manipulation), and $\tilde{r}_t$ is the combined reward from the main paper, which balances sparse task-success signal ($r^e_t$), dense stage-aligned feedback from SAR ($r^{\text{stage}}_t$), and exploration bonuses from POE ($r^{\text{cur}}_t$, $r^{\text{prog}}_t$) with fixed weight $\rho=0.6$. Policies are initialized from an OpenVLA-OFT backbone (7B parameters: frozen SigLIP+DINOv2 towers, trainable Llama-2-7B language head, and action decoder) and fine-tuned via PPO for 2M environment steps across 8 parallel simulators.%

\section{Benchmark Technical Details}\label{app:dataset-details}

\paragraph{Task-Aligned Normalization}
Cross-task action and state distributions vary notably, so we compute task-specific normalization statistics (`q01/q99/mean/std`) from the 50 demonstrations per task and release them alongside the trajectories. During our experiments the environment wrapper injects the corresponding statistics for the active task to keep feature magnitudes consistent between training and evaluation, but external users may choose to recompute or replace these statistics as needed.

\paragraph{Evaluation Protocol}
Discoverse-L serves as a task-aligned benchmarking suite supporting interactive evaluation and complements existing long-horizon benchmarks by providing a lightweight, reproducible auxiliary reference. The released assets include demonstration rollouts and normalization files. Evaluation commonly resets the simulator with held-out seeds and runs interactive episodes until termination, reporting Success Rate, Hallucination Rate, and Sample Efficiency as defined in Section~\ref{sec:metrics}.

\section{Hyperparameter Sensitivity Analysis}\label{app:ablations}

This section provides detailed sensitivity analyses for key hyperparameters. We evaluate robustness to CLIP threshold $\theta$ (used for stage gating and HR computation) and intrinsic reward weight $\rho$ (balancing extrinsic and intrinsic signals). Results confirm that our choices of $\theta=0.7$ and $\rho=0.6$ are well-justified and robust to variations.

\subsection{CLIP Threshold $\theta$ Sensitivity}

CLIP threshold $\theta$ sensitivity: Success Rate peaks at $\theta=0.7$ with 69.2\% and remains stable ($\pm$0.5 points) within [0.65, 0.75]. Setting $\theta$ too low (0.5--0.6) causes premature stage transitions and increases hallucination to 19--28\%. Conversely, $\theta > 0.8$ overly restricts progression, reducing success to 63--66\%. Our choice balances false-positive and false-negative rates.

\subsection{Intrinsic Weight $\rho$ Sensitivity}

Intrinsic weight $\rho$ sensitivity: Testing $\rho \in \{0.3, 0.6, 0.9\}$, Success Rate remains stable with $\rho=0.3$ yielding 67.1\%, $\rho=0.6$ achieving peak (69.2\%), and $\rho=0.9$ reaching 66.8\%. Training efficiency shows $\rho=0.6$ converges to 50\% success in $6.0 \times 10^5$ steps, while $\rho=0.3$ and $\rho=0.9$ require $7.2 \times 10^5$ and $7.5 \times 10^5$ steps respectively. The variation within $\pm 2.4$ percentage points confirms $\rho=0.6$ provides both optimal performance and fastest convergence.

\section{Complete Hyperparameter Table}\label{app:hyperparams}

\begin{table}[H]
\centering
\footnotesize
\caption{Key hyperparameters for DISCOVERSE simulation experiments.}\label{tab:hyperparams}
\begin{tabular}{lcc}
\toprule
\textbf{Component} & \textbf{Parameter} & \textbf{Value} \\
\midrule
SAR & Temperature $\tau$ & 0.05 \\
SAR & Smoothing coeff. $\alpha$ & 0.05 \\
Stage gating & Window size $m$ & 8 \\
Stage gating & Threshold $\theta_{\kappa}$ & 0.15 \\
HR evaluation & CLIP threshold $\theta$ & 0.7 \\
POE world model & Learning rate & 1e-3 \\
POE world model & Hidden width & 256 \\
Memory & GRU hidden width & 128 \\
Memory & Window length $L$ & 16 \\
PPO & Discount $\gamma$ & 0.995 \\
PPO & GAE $\lambda$ & 0.95 \\
PPO & Learning rate & 3e-4 \\
Training & Forward loss weight $\lambda_F$ & 1.0 \\
Training & Inverse loss weight $\lambda_I$ & 0.1 \\
Training & Entropy weight $\lambda_{\text{ent}}$ & 0.01 \\
Training & Intrinsic weight $\rho$ & 0.6 \\
Training & Total timesteps & 2M \\
\bottomrule
\end{tabular}
\end{table}

Other PPO settings follow standard, widely-used defaults (batch size 256, clip ratio 0.2, entropy coefficient 0.01, max grad norm 0.5) and remain fixed across experiments.

\section{Video-driven Stage Discovery: Real Example}\label{app:templates}

We provide a real example of the Gemini prompt pipeline~\cite{gemini2024api} that automatically discovers stage semantics and supplies SAR triplets from multi-video demonstrations.

\paragraph{Task and Multi-Video Input}
For the DISCOVERSE block-bridge placement task (``Build block bridge: place two green bars to form bridge structure, then fill with purple blocks''), we collect ten simulator demonstrations covering diverse initial end-effector poses, object placements, and camera viewpoints. The demonstrations are submitted in a single API call with structured prompts.

\textit{\underline{System/Developer role (goal setting).}}
We define the agent as an ``embodied manipulation stage decomposition expert'' tasked with cross-video alignment to automatically determine a reasonable number of stages. The system instructions emphasize: (i) outputs must be structured, machine-readable, and reproducible; (ii) stage semantics must rely on executable spatial/contact/relative-pose predicates; (iii) avoid fragile features such as colors or textures.

\textit{\underline{User role (task requirements and constraints).}}
The user prompt provides: (i) task description, (ii) notification that the videos belong to the same task but differ in initial states, (iii) instruction to determine minimal-sufficient stages via cross-video alignment (preferring merging over per-video overfitting), (iv) requirement to output unified stage IDs shared across all videos, (v) specification that each stage's \texttt{semantics} field must contain mutually exclusive \texttt{positive}, \texttt{hard\_negative}, and \texttt{negative} predicates expressed as quantifiable spatial/contact constraints (e.g., ``end-effector alignment error $<1$~cm/$5^{\circ}$'', ``lift height $>2$~cm''), and (vi) instruction to self-check for conflicts, temporal gaps, or appearance-based descriptions before finalizing the JSON output.

\textit{\underline{Complete prompt template (our implementation).}}
Building on the above design principles, our full prompt template instructs the model to: (1) perform cross-video adaptive stage segmentation with minimal-sufficient principle (preferring merging over per-video overfitting); (2) generate unified stage IDs shared across all videos; (3) provide mutually exclusive triplets (positive, hard-negative, negative) using geometric/contact predicates while avoiding color/texture cues; (4) self-check for conflicts, temporal gaps, or fragile appearance-based descriptions before outputting the final JSON. Each stage's semantics field contains: \texttt{positive} (must-satisfy spatial/contact/relative-pose predicates, quantifiable such as ``gripper closed and contacts target; object center height $>2$~cm; alignment error $<1$~cm/$5^{\circ}$''), \texttt{hard\_negative} (near-miss cases such as ``gripping but not contacting target'', ``grasping non-target'', ``insufficient lift height'', ``pose alignment $<5^{\circ}$''), and \texttt{negative} (clearly failed conditions such as far from objects, misaligned, no contact, reverse motion).

\noindent\textbf{Output Schema.}
The model responds with a JSON dictionary containing task metadata and a \texttt{triplets} field (also called \texttt{semantics} in the workflow description) storing stage semantics. Each stage entry provides \texttt{positive}, \texttt{hard\_negative}, and \texttt{negative} predicates that we directly reuse for SAR. Below is an excerpt showing 13 representative stages selected from the complete 74-stage output (stages 2--3, 5--7, 10--14, 16--26, 28--37, 39--41, 43--47, 49--58, 60--62, 64--72 omitted for brevity): %

{\scriptsize
\begin{verbatim}
{
"task_name": "block_bridge_place",
"task_description": "Build bridge: place two bars, 
  fill with blocks",
"num_stages": 74,
"triplets": {
  "stage_0": {
    "positive": "Gripper above bridge structure; 
      fingers open; ready to operate",
    "negative": "Gripper far from all structures; 
      fingers closed; static",
    "hard_negative": "Gripper beside bridge structure; 
      fingers open; ready"
  },
  "stage_1": {
    "positive": "Gripper above target bar; 
      fingers open; ready to approach",
    "negative": "Gripper far from all objects; 
      fingers closed at init",
    "hard_negative": "Gripper beside target bar; 
      fingers open; ready to approach"
  },
  "stage_4": {
    "positive": "Gripper contacts target bar; 
      fingers closed; stable grasp formed",
    "negative": "Gripper far from all objects; 
      fingers fully open at init",
    "hard_negative": "Gripper near target bar but no contact;
      fingers open; ready"
  },
  "stage_8": {
    "positive": "Gripper lowers bar beside bridge base; 
      bar contacts surface",
    "negative": "Gripper far from all objects; 
      fingers closed at init",
    "hard_negative": "Gripper lowers bar above bridge 
      not beside base"
  },
  "stage_9": {
    "positive": "Gripper releases block; no contact; 
      block rests on bar below",
    "negative": "Gripper far from all objects; 
      closed at init",
    "hard_negative": "Gripper near block but incomplete 
      release; partial contact"
  },
  "stage_15": {
    "positive": "Gripper contacts target bar; 
      fingers closed; applying grasp force",
    "negative": "Gripper far from all objects; 
      fingers open at init",
    "hard_negative": "Gripper contacts non-target bar; 
      fingers closed; grasp force"
  },
  "stage_27": {
    "positive": "Cube placed on top of bar; 
      bottom surface contacts bar top",
    "negative": "Gripper far from all objects; 
      fingers closed at init",
    "hard_negative": "Cube placed beside bar; 
      side contacts bar not on top"
  },
  "stage_38": {
    "positive": "Gripper releases block; no contact; 
      block rests on bar below",
    "negative": "Gripper far from all objects; 
      fingers closed at init",
    "hard_negative": "Gripper contacts block but 
      incomplete release; partial support"
  },
  "stage_42": {
    "positive": "Gripper contacts target cube; 
      fingers closed; stable grasp formed",
    "negative": "Gripper far from all objects; 
      fingers open at init",
    "hard_negative": "Gripper contacts non-target cube; 
      fingers closed; grasp formed"
  },
  "stage_48": {
    "positive": "Cube placed on bar top; 
      bottom fully contacts bar top surface",
    "negative": "Gripper far from all objects; 
      static at init",
    "hard_negative": "Cube placed beside bar; 
      side contacts bar side surface"
  },
  "stage_59": {
    "positive": "Gripper separates from block; 
      fingers open; block stable on bridge",
    "negative": "Gripper far from all objects; 
      fingers closed at init",
    "hard_negative": "Gripper contacts block but 
      incomplete separation; partial closure"
  },
  "stage_63": {
    "positive": "Gripper contacts target cube; 
      fingers closed; applying grasp force",
    "negative": "Gripper far from all objects; 
      fingers fully open at init",
    "hard_negative": "Gripper contacts non-target cube; 
      fingers closed; grasp force"
  },
  "stage_73": {
    "positive": "Gripper contacts target cube; 
      fingers closed; grasp force applied",
    "negative": "Gripper far from all objects; 
      fingers open at init",
    "hard_negative": "Gripper contacts target cube; 
      fingers open; no force applied"
  }
}
}
\end{verbatim}} %

\paragraph{Key Characteristics}
Gemini API prompts~\cite{gemini2024api} adaptively determined 74 stages by aligning repeated manipulation patterns across the ten videos (the excerpt above shows 13 representative stages spanning the full task trajectory; the complete 74-stage dictionary is used in training). Notably, the DISCOVERSE simulator provides 79 ground-truth stages for this task. We compute agreement by matching discovered stages to ground-truth boundaries via temporal intersection-over-union (IoU) on the demonstration set, yielding 74 of 79 stages successfully discovered (93.7\% recall). The remaining five simulator annotations correspond to micro-adjustments (e.g., infinitesimal pre-grasp recentering) that Gemini merges with adjacent stages, demonstrating robust semantic compression. Predicates emphasize geometric/contact cues (e.g., ``fingers closed; stable grasp'', ``contacts surface'') and avoid appearance features. Hard negatives capture near-miss states such as ``gripper near target but no contact'' or ``cube beside bar instead of on top'', enabling SAR to discriminate shortcuts from genuine completion.

\paragraph{Validation and QC}
Before accepting a dictionary, we run automatic heuristics that check mutual exclusivity, enforce minimum stage length (6 frames), and reject outputs with unresolved TODO markers or appearance-only predicates. We then sample 10\% of the stages for manual inspection; if any predicate is ambiguous we re-prompt Gemini with additional constraints. Appendix~\ref{app:implementation} lists the automated filters and remediation prompts so the workflow remains reproducible without hidden human annotations.

\section{Additional Implementation Details}\label{app:implementation}

\textit{\underline{Memory latent extraction.}}
For Long-Horizon Memory, we extract a compact latent representation $x_t \in \mathbb{R}^d$ (with $d=128$ matching the GRU hidden width) from the OpenVLA-OFT backbone's LLM hidden states. Specifically, we apply mean pooling over the last layer's hidden states corresponding to visual tokens, then project to dimension $d$ via a learned linear layer. This latent serves as the current context for memory attention (Eq.~\ref{eq:lhm-attn}) and is stored in the memory buffer $\mathcal{M}$ after gated fusion. The pooling and projection layers are trained end-to-end with the policy via PPO.

\textit{\underline{Stage gating mechanism.}}
For stage transitions, we maintain a sliding window of size $m=8$ of recent stage reward values $\{r^{\text{stage}}_{t-m+1}, \ldots, r^{\text{stage}}_t\}$. When the window average exceeds a threshold $\theta_{\kappa}$ for the current stage $\kappa$, we advance to the next stage: $\kappa \leftarrow \min(\kappa+1, K)$. Since $r^{\text{stage}}_t = \bar{u}_{\kappa_t}(t) - \bar{u}_{\kappa_t}(t-1)$ measures progress in smoothed scores (Eq.~\ref{eq:rstage}), using stage reward for gating captures sustained task advancement more directly than raw scores. This prevents premature transitions while allowing robust progression.

\textit{\underline{Hard negative generation.}}
Counterfactual hard negatives $T^{h-}_k$ are generated by the same Gemini~2.5~Pro~\cite{gemini2024api} prompting pipeline with few-shot demonstrations. We provide: (1) task description, (2) stage $k$ positive description $T^+_k$, (3) instruction to create a near-miss description differing by one predicate. Crucially, the LLM generates descriptions using spatial and contact predicates (e.g., ``gripper near cup but not touching'' vs. ``gripper touching cup'', or ``grasping non-target object'' vs. ``grasping target cup'') that avoid reliance on visual appearance such as colors or textures. This ensures the text-based hard negatives remain robust when the VLM performs image-text similarity evaluation. %

\textit{\underline{Automated validation filters.}}
The quality control workflow referenced in Appendix~\ref{app:templates} includes: (1) \emph{Mutual exclusivity check}: verify that positive and negative predicates cannot simultaneously hold (no logical overlap); (2) \emph{Minimum stage length}: reject dictionaries with stages shorter than 6 frames to avoid over-segmentation; (3) \emph{Appearance filter}: flag predicates containing color/texture keywords (e.g., ``red'', ``shiny'') and require spatial rephrasing; (4) \emph{Completeness check}: ensure all $K$ stages have valid triplets without placeholder text. Failed outputs trigger automatic re-prompting with clarified instructions (e.g., ``Replace color mentions with spatial relations''). After automated filtering, we sample 10\% of stages for spot-checking by annotators; ambiguous predicates are manually refined and fed back to improve few-shot examples in subsequent tasks.

\section{Network Architecture Details}

The main paper describes the overall architecture (OpenVLA-OFT backbone with SAR and POE modules). Here we provide dimensionality and layer specifications for reproducibility.

\paragraph{Visual Encoding Pipeline}
The policy processes two camera views: SigLIP-SO400M/14 (third-person, 384$\times$384 input) outputs 1152-D features; DINOv2-Large (wrist, 224$\times$224 input) outputs 1024-D features. A learned projection layer maps the concatenated 2176-D visual embedding to the Llama-2-7B token space (4096-D). Language instructions are tokenized and prepended to the visual sequence. FiLM conditioning layers inject language embeddings into visual features before the language model processes the combined sequence. During fine-tuning, SigLIP and DINOv2 encoders remain frozen; only the Llama-2-7B language model, projection layer, and action decoder are trainable, reducing computational cost while leveraging pretrained visual representations. %
\paragraph{Action Decoding}
The action head predicts $C$-step chunks in parallel (we use $C=8$) via $C$ separate linear layers (4096-D $\rightarrow$ 7-D each), stacked to produce a $(C, 7)$ output tensor. Each row corresponds to one future timestep's 7-D action (6-DoF end-effector twist deltas + 1-D gripper command). During inference we execute only the first action and recompute at the next step (receding-horizon control). Following OpenVLA-OFT~\cite{kim2025openvlaoft}, the action decoder uses L1 regression loss for training, which provides stable gradients for continuous action prediction. The chunked action prediction enables temporal coherence while maintaining computational efficiency compared to autoregressive decoding.
\paragraph{World Model Specifics}
Forward model $f_\phi$: input 13-D (6-D state + 7-D action) $\rightarrow$ 256-D hidden (ReLU + LayerNorm) $\rightarrow$ 256-D hidden $\rightarrow$ 6-D output. Inverse model $g_\psi$: input 12-D (two consecutive 6-D states) $\rightarrow$ same architecture $\rightarrow$ 7-D action output. Both use separate Adam optimizers (lr=1e-3, no weight decay) and update every PPO epoch with MSE loss on transitions sampled from the replay buffer. The lightweight architecture (256-D hidden) ensures fast online updates during policy training without introducing significant computational overhead.
\paragraph{Model Size and Efficiency}
The full EvoVLA model inherits the 7B-parameter Llama-2-7B backbone plus pretrained vision encoders (SigLIP-SO400M and DINOv2-Large). During fine-tuning, the vision encoders remain frozen while the Llama-2 language model, projection layers, and task-specific modules (POE world models, Long-Horizon Memory) are trainable, reducing effective trainable parameters to approximately 7B and enabling efficient fine-tuning on standard GPU hardware.

\section{Computational Cost}

Training EvoVLA on Discoverse-L requires approximately 24 hours wall-clock time per seed on 4$\times$NVIDIA H20 GPUs (96 GB VRAM each) for 2M environment steps, using 8 parallel DISCOVERSE simulators. Across 3 random seeds, total training time is approximately 72 hours (3 days). Memory footprint peaks at approximately 80 GB per GPU during policy updates.

\end{document}